\documentclass[11pt]{article}
\usepackage[T1]{fontenc}
\usepackage[utf8]{inputenc}
\usepackage{arxiv_card}
\usepackage{amsmath,amssymb,booktabs,tabularx,array,graphicx}
\usepackage{hyperref}
\hypersetup{colorlinks=true,linkcolor=cardaccent,citecolor=cardaccent,urlcolor=cardaccent}
\usepackage{multirow}
\usepackage{url}
\usepackage{wasysym}
\usepackage{enumitem}
\newcommand{\benchmark}{\textsc{Ego-SpaCR}}
\newcommand{\question}{Spatially grounded Conversational Reasoning}
\newcommand{\questionshort}{\textsc{SpaCR}}
\newcommand{\method}{Spatially grounded Conversational Memory}
\newcommand{\methodshort}{\textsc{SpaC-MEM}}

\newcolumntype{L}[1]{>{\raggedright\arraybackslash}p{#1}}
\newcolumntype{Y}{>{\raggedright\arraybackslash}X}

\title{Have I Scene This Before? Spatially Grounded Conversational Memory for Complex Queries in Egocentric Assistants}
\runningtitle{Spatially Grounded Conversational Memory for Egocentric Assistants}
\author{Jiazhou Liang$^{1*}$, Liam Gallagher$^{1*}$, Kiko Chen$^{1*}$, David Guo$^{1*}$, Armin Toroghi$^{1}$, Yifan Simon Liu$^{1}$, Scott Sanner$^{1}$}
\hypersetup{
  pdftitle={Have I Scene This Before? Spatially Grounded Conversational Memory for Complex Queries in Egocentric Assistants},
  pdfauthor={Jiazhou Liang, Liam Gallagher, Kiko Chen, David Guo, Armin Toroghi, Yifan Simon Liu, Scott Sanner}
}
\affiliation{$^{1}$University of Toronto \\ $^{*}$Equal contribution}
\date{\today}
\logo[1.1cm]{figures/logo.pdf}

\begin{document}
\begin{abstract}

Personal assistants are evolving from search and text interfaces toward interactive multimodal systems that retain spatial and conversational history to support ongoing tasks. Effective assistants must therefore reason jointly over observed scenes and past dialogue to address complex requests. We formalize this challenge as \emph{\question{}} (\questionshort{}), covering cross-scene, recall-oriented, and counterfactual spatial queries that require accurate geometric reasoning and the integration of dialogue with spatial evidence.
Existing approaches are poorly suited to this setting. 
Direct VLMs incur high inference costs and context limits as multimodal histories grow. Keyframe selection and retrieval can omit evidence needed for complete recall, while memory systems must preserve both conversational facts and object geometry to support spatial reasoning.
To bridge this gap, we propose \method{} (\methodshort{}), an object-centric working memory that applies 3D reconstruction and segmentation to egocentric observations, thereby spatially
grounding conversational context directly to physical objects. It compresses long multimodal histories while preserving spatial and conversational evidence and allowing object-specific information to be updated through later dialogue.
We further introduce \benchmark{}, an egocentric \questionshort{} dataset comprising 620 video sessions from ScanNet, augmented with 95 conversations averaging 230 dialogue turns each, and 3,100 evaluation queries.
Experiments show that \methodshort{} achieves the highest overall answer accuracy among the evaluated methods and improves object recall, while requiring substantially fewer reference input tokens than native-video baselines. Analysis shows that removing 3D spatial information substantially degrades performance, highlighting the importance of preserving both spatial and conversational evidence for an egocentric assistant.

\end{abstract}

\maketitle
\section{Introduction}
\label{sec:intro}

\begin{figure}[t]
\centering
\includegraphics[width=\linewidth]{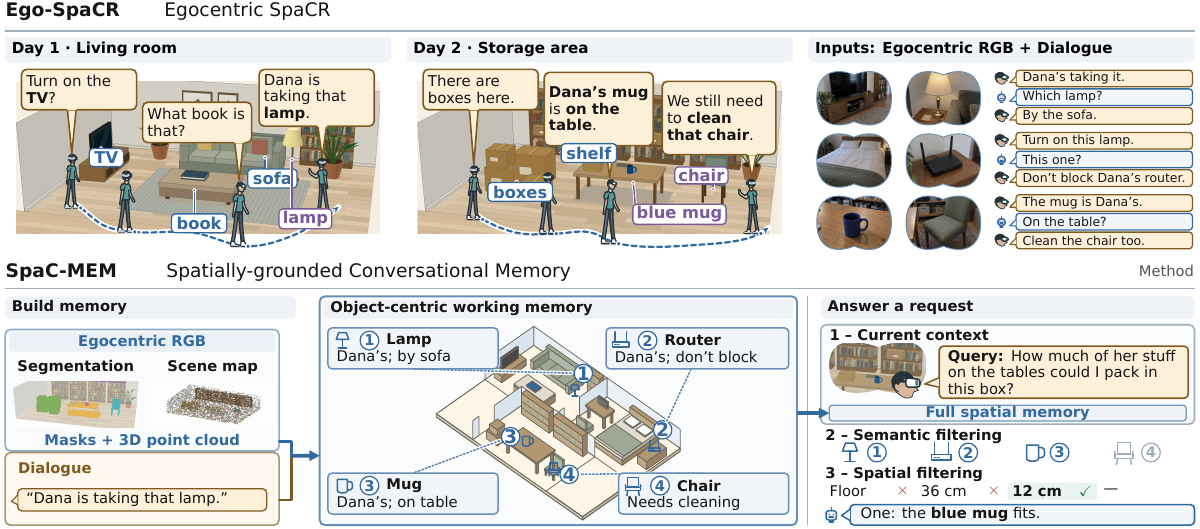}
\caption{\questionshort{} involves complex spatial queries that require connecting user-stated facts to spatial evidence about the same physical objects across the interaction history. Here, identifying which of Dana's belongings could fit in a box requires recalling ownership established through conversation and checking previously observed locations and object dimensions. \methodshort{} links this conversational and spatial evidence through persistent objects in a unified working memory.}

\label{fig:teaser}
\end{figure}

As personal assistants move beyond search and text interfaces toward egocentric multimodal systems, they must remain coherent with both what the user sees and what they say over time \citep{wang2023holoassist,liang-etal-2026-evaluating}. Users naturally refer to objects in view while expressing facts such as ownership and preferences that cannot be inferred visually. As interactions span rooms, sessions, and days, tasks requiring object comparisons, alternative arrangements, or a complete inventory arise, even when the relevant evidence is no longer visible or recoverable. Thus, an effective assistant should preserve both spatial and conversational context to support ongoing tasks, without requiring the user to remember or revisit earlier interactions.
Figure~\ref{fig:teaser} demonstrates how a user might identify friends' belongings while moving through an apartment, then later ask a spatially grounded question dependent on those references, such as \textit{``How much of my friend's stuff on the tables earlier can I fit in this box?''}
Answering such complex cross-scene spatial queries requires more than retrieving a relevant frame: the assistant must recover conversational facts tied to persistent objects, combine them with prior geometric, qualitative relational, or counterfactual spatial evidence, and account for all relevant objects across the history. We refer to this as \emph{\question{} (\questionshort{})}.


\textbf{Where Existing Approaches Fall Short.} Direct vision-language models (VLMs) repeatedly process growing multimodal histories, exhausting context windows and incurring high inference costs \citep{tang2025aks,zhu2026focus}. Keyframe selection and multimodal Retrieval-Augmented Generation (RAG) reduce context but may miss objects or evidence required for questions requiring complete recall. Spatial scene memories primarily retain visual observations without evolving conversational context \citep{gu2024conceptgraphs,yang2025_3dmem}. Multimodal conversational memories such as M2A retain raw interactions alongside semantic records \citep{feng2026m2a}, but do not maintain the explicit object geometry needed for complex spatial reasoning. Existing datasets likewise lack key requirements of \questionshort{} (Table~\ref{tab:related}). VideoQA approaches~\citep{yang2025thinking,yang2025cambrian,gu2026wild} do not evaluate spatial reasoning over conversational facts, while situated dialogue datasets~\citep{wu2023simmc} focus on single-scene interactions with limited spatial reasoning.

\textbf{A Spatially Grounded Working Memory.} To address this gap, we propose \method{} (\methodshort{}), a unified object-centric \emph{working memory} that grounds user-specific conversational information to physical objects in historical egocentric interactions while preserving their spatial evidence (e.g., object size) via 3D geometric reconstruction with segmentation. \methodshort{} compresses long frame-based histories into a compact representation, while allowing conversational information to be accumulated and updated over time (e.g., changing a chair from discarded to kept).
We further introduce \benchmark{}, a benchmark for \questionshort{} over egocentric video, augmented with task-oriented dialogue. Our experiments show that \methodshort{} achieves the highest overall answer accuracy among the evaluated methods and improves object recall while requiring substantially fewer reference input tokens than native-video baselines. Removing explicit 3D spatial information substantially degrades performance, supporting the use of spatial evidence alongside conversation in memory.
Our contributions are summarized as follows:
\begin{itemize}
\item We introduce \method{} (\methodshort{}), an object-centric working memory that connects detailed spatial and conversational information via persistent physical objects across historical egocentric user interactions.

\item We introduce \benchmark{}, a benchmark for \questionshort{} over egocentric video and task-oriented dialogue, with 95 task-driven conversations in 620 real-world video sessions and 3,100 evaluation questions across varying classes of spatial complexity and recall breadth.

\item We conduct a controlled analysis across \benchmark{}, revealing where existing methods fail as queries require increasingly complex spatial reasoning and recall over object sets.

\item We show that \methodshort{} consistently outperforms existing methods in accuracy and object recall, while substantially reducing input token cost compared to in-context methods.

\end{itemize}

\section{Related Work}
\label{sec:related}

\subsection{Egocentric Video and Dialogue Benchmarks}

\emph{Egocentric video} captures the environment from the wearer's viewpoint, recording what they see and interact with \citep{damen2018epic,grauman2022ego4d}. EPIC-KITCHENS \citep{damen2018epic} and Ego4D \citep{grauman2022ego4d} capture longer-term first-person experiences across different environments. ScanNet further provides RGB-D observations of real scenes together with geometry and object annotations \citep{dai2017scannet} as the foundation of spatial understanding evaluation.

\textbf{Video question answering (VideoQA).}
Egocentric VideoQA benchmarks, such as EgoVQA and EgoSchema, cover first-person activities over longer video contexts \citep{fan2019egovqa,mangalam2023egoschema}. OpenEQA and SQA3D evaluate questions in situated 3D environments \citep{majumdar2024openeqa,ma2023sqa3d}; VSI-Bench, VSI-SUPER, and VSI-Super-Wild focus on spatial reasoning, long-term spatial recall, and counting \citep{yang2025thinking,yang2025cambrian,gu2026wild}.
These VideoQA benchmarks evaluate questions about observations without an evolving history of user-stated facts tied to persistent objects. In \questionshort{}, language is integral to the evolving history, supplying facts by referring to an object in view, such as ownership or preferences. The assistant must connect those facts to the object's spatial evidence over time.

\textbf{Long-term Dialogue and Situated Conversation.}
Long-term dialogue benchmarks such as LongMemEval~\citep{wu2025longmemeval} evaluate assistants' ability to use evolving conversation but do not ground it in spatial environments. Situated dialogue (SIMMC-VR~\citep{wu2023simmc} and SIMMC 2.0~\citep{kottur2021simmc}) connect conversation with spatial properties in simulated environments, but focus on short, single-scene interactions with limited spatial complexity. They do not require reasoning across scene spaces to answer the complex counterfactual spatial queries that users may request. Related recommendation work studies visually grounded fashion dialogue and scene-based in-situ item labeling \citep{guo2026vogue,liang-etal-2026-evaluating}, connecting user preferences with visual item information.

\subsection{Spatial Understanding with Video Input}


\textbf{In-Context Spatial Understanding.}
Recent multimodal models directly process video and language; Gemini supports native video understanding, while Qwen3-VL extends long-context vision-language modeling to long videos and temporal grounding \citep{gemini37flashcard,bai2025qwen3}. These models keep visual and conversational history in context instead of organizing it around persistent physical objects. As a result, recovering every object linked to conversational facts remains difficult.
Query-aware frame selection methods such as AKS and FOCUS select informative frames before VLM inference \citep{tang2025aks,zhu2026focus}. But selection is conditioned primarily on queries and visuals, without considering the evolving conversation.

\textbf{Agentic Spatial Understanding with Object Memory.}
Object memory methods reconstruct scenes from observations. ConceptGraphs builds open-vocabulary 3D scene graphs from RGB-D input \citep{gu2024conceptgraphs}. However, they primarily target visual understanding. Agentic systems (e.g., VideoAgent) combine captions, object tracking, re-identification, and tool-based memory in agentic retrieval \citep{fan2024videoagent}; Gemini's agentic video understanding searches and inspects segments. While these systems combine language-guided reasoning with video memory, they still keep conversation and visual memory parallel.

\textbf{Multimodal RAG and Dialogue Memory.}
Retrieval-based methods select stored evidence for answering: RAG retrieves text \citep{lewis2020rag}, with related dense retrieval work studying manifold-aware distances and LLM-guided Bayesian active learning \citep{liu-etal-2025-ma-dpr,kim-etal-2026-bayesian}; natural language recommendation also uses LLM relevance judgments to model item relevance \citep{liu-etal-2026-multimodal}. VideoRAG retrieves video evidence \citep{jeong2025videorag}. Our \emph{Multimodal RAG} baseline retrieves dialogue and frames using embeddings, while \emph{Summary RAG} retrieves textual summaries of both. M2A retains raw interactions alongside semantic memory and uses agentic retrieval to manage long-term context \citep{feng2026m2a}. Related conversational memory methods organize access through tree-structured semantic queries, preserve temporal order with segment trees, or decompose goals for targeted retrieval \citep{liu-etal-2026-semantic,liu2026segtreemem,liang2026goalmem}. Retrieval can miss evidence needed for complete recall, while summaries can omit geometry required for \questionshort{}.

\section{\questionshort{}: Spatially Grounded Conversational Reasoning}
\label{sec:questions}

\begin{figure}[tbp]
    \centering
    \includegraphics[width=\linewidth]{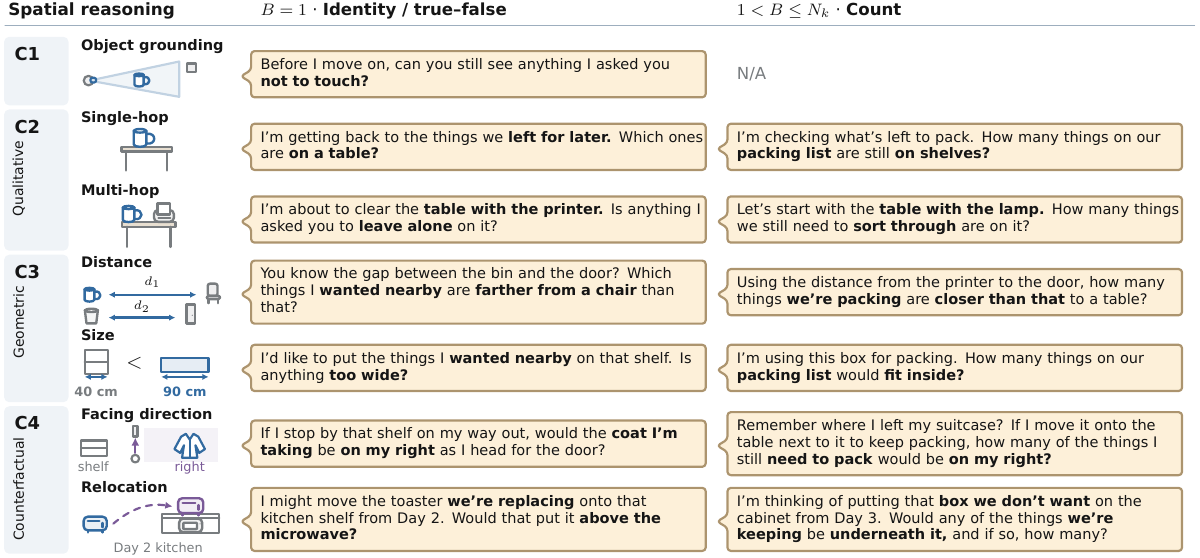}
\caption{Example \questionshort{} queries by recall type (columns) and reasoning level (rows). Unlike VideoQA, each query requires user-stated facts from conversation alongside spatial evidence.}
    \label{fig:complexity}
\end{figure} 


We distinguish \questionshort{} from existing settings through four required properties: (i) \emph{bimodal histories}, since users naturally introduce non-visual information while referring to objects they see; (ii) \emph{spatial grounding}, so that this information remains attached to the same physical objects across views, sessions, and revisits; (iii) \emph{complex spatial reasoning}, because later requests may depend on geometric relations or hypothetical arrangements that are not directly visible; and (iv) \emph{recall breadth}, as tasks may require accounting for the complete relevant object set rather than retrieving isolated evidence.

\textbf{(i) Bimodal histories.}
An egocentric assistant receives two histories: a sequence of RGB observations $\mathcal{V}=(v_1,\ldots,v_T)$ and a dialogue history $\mathcal{C}=(c_1,\ldots,c_K)$ spanning multiple scenes.
Each frame $v_t$ contains a set of physical objects $\mathcal{O}_t$, and each turn $c_k$ occurs at time $t_k$ with $v_{t_k}$. The objects observed by turn $k$ are $\mathcal{O}_{\leq t_k}=\bigcup_{t\leq t_k}\mathcal{O}_t$. Object identities persist within a scene; objects from different scenes remain distinct. We represent each user-stated semantic fact as $\phi=(o,p,x)$, where object $o$ has property $p$ with value $x$, such as $(o,\mathrm{owner},\mathrm{Dana})$, capturing information unavailable from $\mathcal{V}$. As users change decisions over time, an $(o,p)$ may change: a \emph{correction} retracts the old value and applies a new one retroactively, while a \emph{state change} preserves it until the update. If a lamp changes from \emph{keep} to \emph{donate}, a query about its initial status (e.g., \textit{``What did I first decide about the lamp?''}) resolves to \emph{donate} under a correction but \emph{keep} under a state change.


\textbf{(ii) Spatially grounded queries.}
Users may ask for help at any point during a task, based on what they previously said and encountered. An egocentric assistant must therefore answer from the interaction history. We define a user request at turn $k$ as a query $q_k$ that can use only $\mathcal{V}_{\leq t_k}$ and $\mathcal{C}_{<k}$. Let $\Phi_{<k}$ contain the facts established before $q_k$, with corrections and state changes resolved for the time the query refers to. Each query defines a conversational condition $h_k(o)$ and a spatial condition $r_k(o)$, giving the matching set $\mathcal{S}_k=\{o\in\mathcal{O}_{\leq t_k}:h_k(o)\land r_k(o)\}$. Depending on the user's need, the assistant may identify an object from $\mathcal{S}_k$ or account for the set through $|\mathcal{S}_k|$, as in tasks such as packing or organizing. This yields a two-dimensional space defined by the \emph{spatial complexity} required by $r_k(o)$ and the \emph{recall breadth} needed to recover $\mathcal{S}_k$ (\autoref{fig:complexity}).


\textbf{(iii) Spatial complexity.}
Users may ask whether a prior object is currently visible, whether objects fit geometrically, are closer to one another, or satisfy hypothetical spatial arrangements. Supporting such requests requires increasingly complex spatial evidence. We define four levels through $r_k(o)$:

\begin{description}[leftmargin=1em,font=\normalfont]
    \item[(C1) \emph{Object grounding}]
     tests whether the assistant can connect a conversational condition $h_k(o)$ established earlier to the same physical object in the current view, with
    $r_k(o)=\mathbf{1}[o\in\mathcal{O}_{t_k}]$.
    Users may refer back to previously discussed objects as they reappear during an ongoing task.

    \item [(C2) \emph{Qualitative relations}]
    tests whether the assistant can evaluate single- or multi-hop spatial relations $R$, such as \emph{on} or \emph{near}, over conversationally grounded objects. For objects on the same table as a conversationally qualified lamp,
    $
    r_k(o)=\exists u,w\in\mathcal{O}_{\leq t_k}:(u:\mathrm{table})\land(w:\mathrm{lamp})
    \land g_k(w)\land R_{\mathrm{on}}(o,u)\land R_{\mathrm{on}}(w,u).
    $
    Here $g_k$ qualifies the anchor independently of the target condition $h_k$: for \emph{``a lamp that my roommate owns,''}
    $g_k(w)=\mathbf{1}[(w,\mathrm{owner},\mathrm{Roommate})\in\Phi_{<k}]$;
    without an anchor qualification, $g_k(w)=1$.

    \item [(C3) \emph{Geometric properties}]
    tests whether the assistant can accurately recover the geometric properties of objects and relations, such as size or distance. Users may ask whether objects fit or compare their distances for tasks like packing, where qualitative relations alone are insufficient.

    \item [(C4) \emph{Counterfactual spatial reasoning}]
    tests whether the assistant can evaluate spatial relations under a hypothetical viewpoint or placement. Answering may require combining geometric evidence from earlier observations to reason about an arrangement that was not directly observed.
\end{description}

\textbf{(iv) Recall breadth.}
Let $\mathcal{R}_k$ denote the target objects supporting the answer, excluding anchors used only to evaluate spatial conditions. For a count query, $\mathcal{R}_k=\mathcal{S}_k$; an identification or positive existence query needs a single target or witness. We define \emph{recall breadth} as $B_k=|\mathcal{R}_k|$. Counting requires recovering the complete matching set. For zero-count and negative existence answers, $\mathcal{R}_k=\varnothing$, but all candidates must be ruled out; recall breadth therefore measures target-set size.

\section{Object-Centric Spatially Grounded Conversational Memory}
\label{sec:method}



\begin{figure}[t]
    \centering
    \includegraphics[width=\linewidth]{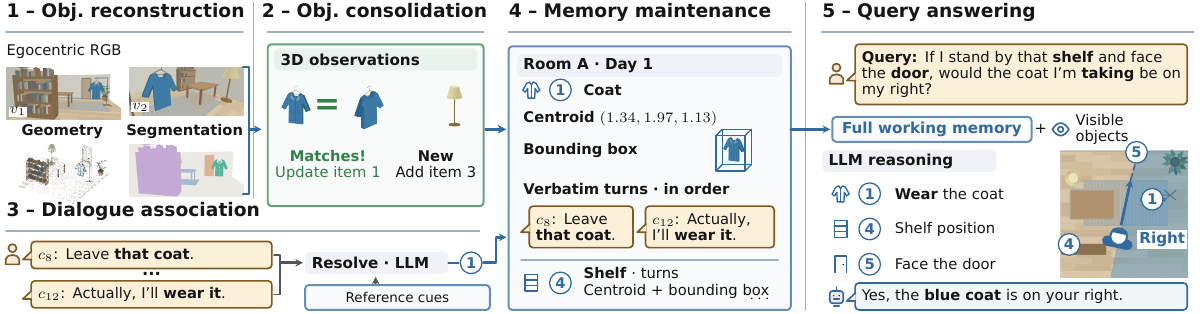}
\caption{\methodshort{} consolidates egocentric observations into persistent 3D object records, associates dialogue turns, and supplies the resulting working memory with visible objects to an LLM. The example combines updated intent and geometry for hypothetical viewpoint reasoning (C4).}

    \label{fig:method}
\end{figure}

As interaction histories grow, assistants that keep the bimodal history in context face difficulty and cost in \questionshort{}. We therefore propose \method{} (\methodshort{}), an object-centric conversational \emph{working memory} that spatially grounds past dialogue in persistent physical objects. Recording each object once keeps the memory far smaller than the history while preserving the spatial and conversational detail \questionshort{} needs. \methodshort{} consists of five stages, illustrated in \autoref{fig:method}: (1) \emph{object reconstruction}, where incoming frames $\mathcal{V}$ are converted into frame-level 3D object observations; (2) \emph{object consolidation}, where repeated observations update existing objects in a persistent pool and newly observed objects are added; (3) \emph{dialogue association}, where incoming turns from $\mathcal{C}$ are grounded to the corresponding objects and attached to their records; (4) \emph{memory maintenance}, which preserves the latest conversational state while retaining the relevant interaction history for each object; and (5) \emph{query answering}, where an LLM reasons directly over \methodshort{} to answer \questionshort{} queries. For full implementation details, see Appendix~\ref{app:method}.

\textbf{(1) Object reconstruction.}
\methodshort{} extracts frame-level object observations $\hat{o}_{t,i}$ and their geometry from the RGB frames at time $t$, $\mathcal{V}_{\leq t}$, without requiring depth measurements or camera poses.
A \emph{feed-forward geometry reconstruction model}, Depth Anything 3~\citep{lin2025depth}, serves as $f_{\mathrm{geo}}$ and estimates the depth map $\hat{D}_t$, the point map giving each pixel's 3D position $\hat{P}_t$, and the camera pose in the environment. \methodshort{} transforms these components into a shared coordinate system, with axes aligned to dominant scene planes. A segmentation model $f_{\mathrm{seg}}$ takes the frame $v_t$ and a predefined object vocabulary $\mathcal{L}$, and returns $n_t$ object instances
$\{(M_{t,i},\ell_{t,i})\}_{i=1}^{n_t}$,
where $M_{t,i}$ is the mask of instance $i$ and $\ell_{t,i}\in\mathcal{L}$ is its class label. \methodshort{} lifts each mask into 3D through the reconstructed point map,
$
\hat{P}_{t,i}=\{\hat{P}_t(u):u\in\tilde{M}_{t,i}\},
$
where $\tilde{M}_{t,i}\subseteq M_{t,i}$ retains only pixels whose depth is consistent with the object, reducing background contamination near mask boundaries. Each nonempty $\hat{P}_{t,i}$ defines an object observation $\hat{o}_{t,i}$ with spatial record
$
s_t(\hat{o}_{t,i})=(\ell_{t,i},\boldsymbol{\mu}_{t,i},\mathbf{B}_{t,i}),
$
containing its class label, 3D centroid $\boldsymbol{\mu}_{t,i}$, and an outlier-robust axis-aligned bounding box $\mathbf{B}_{t,i}$.
A key design choice is whether \methodshort{} should preserve this detailed spatial evidence or retain only object class and persistent identity. We evaluate this choice in RQ3.

\textbf{(2) Object consolidation.}
\methodshort{} consolidates the frame-level observations into a persistent object pool $\hat{\mathcal{O}}_{\leq t}$ for each scene. For each observation $\hat{o}_{t,i}$, we consider previously reconstructed objects in $\hat{\mathcal{O}}_{\leq t-1}$ with the same class label $\ell_{t,i}$ and select
$
a_t(\hat{o}_{t,i})=\arg\max_{\hat{o}} A(\hat{o}_{t,i},\hat{o}),
$
where $A$ measures overlap between their reconstructed 3D regions. The association is accepted only when the best match is unambiguous, reducing erroneous merges between nearby objects. If $\hat{o}_{t,i}$ matches no existing object, it initializes a new persistent object. For a matched object, its accumulated spatial record $s_{\leq t}(\hat{o})$ is updated with the new observation. The pool is periodically refined by merging overlapping fragments that never co-occur in the same frame and are supported by appearance or tracking evidence, while rarely observed objects are removed.
A key design choice is \emph{object-centric} vs. \emph{per-frame} structure that receives the same reconstructed geometry before cross-frame association. Object-centric structures prevent the downstream context from growing with every observed frame while preserving detailed spatial information that exceeds the context window. 

\textbf{(3) Dialogue association.}
After object consolidation establishes persistent object identities, this step grounds each dialogue turn to the physical objects it refers to, allowing conversational information to accumulate on the same object records over time. For a turn $c_k$ at visual time $t_k$, an association model maps
$
\hat{\mathcal{A}}_k=f_{\mathrm{assoc}}
(c_k,\hat{\mathcal{O}}_{t_k},\hat{\mathcal{O}}_{\leq t_k},\mathcal{C}_{<k}),
$
where $\hat{\mathcal{O}}_{t_k}$ is the subset of persistent objects observed at frame $t_k$, and $\hat{\mathcal{A}}_k\subseteq\hat{\mathcal{O}}_{\leq t_k}$ contains the objects about which the turn makes a statement. The model resolves references using prior dialogue, object classes, spatial descriptions, pronouns, and previously established names.

Each $\hat{o}\in\hat{\mathcal{A}}_k$ is then associated with the original turn $c_k$; we retain the original dialogue turn rather than replace it with a generated summary \citep{hu_does_2026}, preserving the user's wording while allowing later turns to revise information attached to the same persistent object. In RQ3, we evaluate the reliability of LLM-predicted object associations with human-annotated references.

\textbf{(4) Memory maintenance.}
Together, object reconstruction, consolidation, and dialogue association produce a persistent working memory that connects each physical object with its accumulated spatial and conversational information. Before query $q_k$, \methodshort{} represents this memory as one record per object $\hat{o}\in\hat{\mathcal{O}}_{\leq t_k}$:
$
\mathcal{W}_k(\hat{o})
=
\left(
s_{\leq t_k}(\hat{o}),
[c_j]_{j<k:\,\hat{o}\in\hat{\mathcal{A}}_j}
\right),
$
where $s_{\leq t_k}(\hat{o})$ is the consolidated spatial record and the list contains all preceding dialogue turns associated with that object. Records are organized by scene and session while preserving conversational order. As new observations and turns arrive, the corresponding record is updated rather than adding separate frame-level entries.

\textbf{(5) Query answering.}
For a query $q_k$, the answering agent $\pi_{\mathrm{LLM}}$ receives the current memory, the objects visible in the frame $\hat{\mathcal{O}}_{t_k}$, and the query $q_k$, producing $\hat{a}_k=\pi_{\mathrm{LLM}}(q_k,\hat{\mathcal{O}}_{t_k}, \mathcal{W}_k)$. The complete working memory is provided without retrieval, allowing the answerer to examine all stored objects when a query requires a complete set rather than a relevant subset. 

\section{Experiments}

To understand the performance and limitations of existing methods in \questionshort{}, we study three RQs\footnote{Code: \url{https://github.com/joeliang0520/SAPCR-MEM}.}:

\noindent
\begin{tabularx}{\linewidth}{@{}>{\bfseries}l X@{}}
RQ1 & Does \methodshort{} improve downstream answering model performance on \questionshort{} compared with existing approaches? \\
RQ2 & Across object identification, geometric estimation, and spatial reconstruction, which abilities limit current methods on \questionshort{}? \\
RQ3 & How do \methodshort{}'s design choices, including object-centric structure, dialogue association, and geometric reconstruction, contribute to overall performance?
\end{tabularx}


We compare against three baseline families spanning direct in-context reasoning, agentic video memory, and retrieval-based conversational memory: 
(i) \emph{In-context spatial understanding.}
Native-video baselines use \textsc{Gemini 3.7 Flash}~\citep{google2026gemini37} and \textsc{Qwen3.8-27B-FP8}~\citep{qwen2026qwen38}, each given the video prefix and timestamped conversation before the query at 1 fps. Frame-based baselines use uniform sampling, AKS, and FOCUS~\citep{tang2025aks,zhu2026focus}, with selected frames capped by a fixed visual-token budget and provided to \textsc{GPT-5.6 Terra}~\citep{openai2026gpt56terra} and \textsc{GLM-5.3 Flash}~\citep{zai2026glm53flash}.
(ii) \emph{Agentic spatial understanding with object memory.}
We evaluate \emph{VideoAgent}~\citep{fan2024videoagent} and the agentic video mode of \textsc{Gemini 3.7 Flash}.
(iii) \emph{Multimodal RAG and conversational memory.}
We evaluate \emph{Multimodal RAG}, \emph{Summary RAG}, and \emph{M2A}~\citep{feng2026m2a}, using \textsc{GPT-5.6 Terra} and \textsc{GLM-5.3 Flash}. We also include Random and Conversation Only controls (Appendix~\ref{app:controls}).




\subsection{Datasets and Metrics}
\label{sec:benchmark}

\begin{figure}[t]
\centering
    \includegraphics[width=1.0\linewidth]{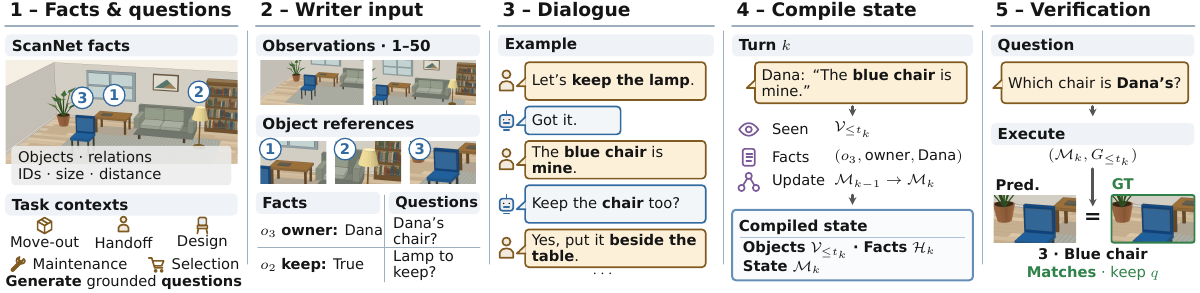}
    \caption{\benchmark{}'s pipeline for natural, verifiable dialogue over egocentric histories.}
    \label{fig:pipeline}
\end{figure}

\textbf{\benchmark{}.} Existing benchmarks do not comprehensively evaluate \questionshort{} (cf. \autoref{sec:related}). We therefore construct \benchmark{}, an egocentric benchmark built from ScanNet~\citep{dai2017scannet} that covers the four properties of \questionshort{} defined in \autoref{sec:questions}. ScanNet provides real egocentric observations with annotated object identities and geometry, enabling verifiable spatial evidence and analysis of reconstruction error. Construction follows five stages (\autoref{fig:pipeline}). 

\begin{description}[leftmargin=1em,font=\normalfont]
\item[(1) \emph{Facts \& questions.}] Five everyday tasks and compatible personas define plausible object-specific semantic facts. Clips from selected ScanNet scenes are composed into multiple visits, forming a shared fact pool $\Phi$. We then construct queries by combining conversational conditions with spatial operations, each represented by a deterministic program $G_k$ that produces a ground-truth answer from corresponding facts and ScanNet geometry. For count queries near geometric boundaries, we allow a bounded answer interval to account for uncertainty in the annotated mesh (cf. \autoref{app:benchmark}).

\item[(2) \emph{Writer input.}] For each scene-level batch, the Writer receives the relevant observations, grounded object references, facts to establish, selected queries, and preceding dialogue context. 

\item[(3) \emph{Dialogue.}] An LLM \emph{Writer} generates a natural continuation that establishes the required facts before each query while maintaining coherence across visits. 

\item[(4) \emph{Compile state.}] An independent LLM \emph{Reader}, without access to the Writer's annotations, processes the conversation turn by turn to reconstruct grounded facts and updates available.

\item[(5) \emph{Verification.}] We retain a query only if all required facts precede $q_k$, all references resolve, and executing $G_k$ over the Reader-recovered state reproduces the target answer, followed by human verification. We evaluate all methods using the following two metrics. 
\end{description}

\begin{table}[t]
\centering
\scriptsize
\caption{Summary of 95 conversations spanning 620 video sessions and 3{,}100 evaluation questions in real ScanNet rooms (mean $\pm$ population s.d.; brackets give [min, max]).}
\label{tab:dataset-stats}
\begin{tabular}{@{}lr@{\hspace{2.5em}}lr@{}}
\toprule
Statistic & Value & Statistic & Value \\
\midrule
Task types & 5 & Grounded objects per conversation & $49.8 \pm 24.3$ \\
Multi-session conversations & 95 & Object classes per conversation & $27.4 \pm 9.8$ (60 total) \\
Video sessions (clips) & 620 (6.4\,h) & Evaluation questions & 3{,}100 \\
Semantic facts per conversation & $201.2 \pm 118.8$ & Questions per conversation & $32.6 \pm 14.9$ \\
Sessions per conversation & $6.5 \pm 2.7$\, [2, 14] & Complexity mix C1\,/\,C2\,/\,C3\,/\,C4 (\%) & 6.0\,/\,24.3\,/\,45.9\,/\,23.8 \\
Video per conversation (min) & $4.1 \pm 1.9$ & Answer type count\,/\,object\,/\,boolean (\%) & 85.4\,/\,13.4\,/\,1.2 \\
Dialogue turns per conversation & $230.1 \pm 96.3$\, [64, 505] & Question length (words) & $34.0 \pm 13.8$ \\
Dialogue words per conversation & $5{,}315 \pm 2{,}534$ & Dialogue turns\,/\,words before a question & $158.8$\,/\,$3{,}797$ \\
\bottomrule
\end{tabular}
\label{tab:stats}
\end{table}


\textbf{Answer accuracy.}
Our primary metric is answer accuracy. For object-identification questions, the model returns a natural-language description of the selected object. We use a VLM only to map this description to a ground-truth object, using reference images, bounding boxes, and spatial information. Accuracy is determined by whether the mapped object matches the ground-truth object (see Appendix~\ref{app:judge} for details).

\textbf{Object recall.}
To verify that a method recovers the objects needed rather than guessing the count, we ask each answer model to also return the object descriptions supporting its count and map them to physical objects using the same VLM. Let $\mathcal{R}_k$ be the ground-truth objects required for count query $q_k$, and $\hat{\mathcal{R}}_k$ the objects returned by the model. For queries with $|\mathcal{R}_k|>0$, object recall is
$
\frac{|\hat{\mathcal{R}}_k \cap \mathcal{R}_k|}{|\mathcal{R}_k|}
$
the fraction of required objects recovered, averaged over nonempty count queries.
We compare the context supplied to the final answer model using reference input tokens under a common scheme across providers (Appendix~\ref{app:token_cost}). For \methodshort{}, we use \textsc{Depth Anything 3}~\citep{lin2025depth} for reconstruction and \textsc{SegVGGT}~\citep{qu2026segvggt} for segmentation.

\subsection{Results}
\label{sec:results}
\begin{table*}[t]
\centering
\scriptsize
\setlength{\tabcolsep}{1pt}
\renewcommand{\arraystretch}{1.18}
\newcommand{\score}[1]{#1}
\newcommand{\bestscore}[1]{\textbf{#1}}
\newcommand{\secondscore}[1]{\underline{#1}}
\newcommand{\na}{\textcolor{gray}{--}}
\providecommand{\SpaC}{SpaC}
\caption{Answer Accuracy across selected baselines and \SpaC-MEM (\%); bold and underlined entries are the best and second-best values per column.}
\label{tab:main-results-no-ci}
\ifdefined\spacPanelL\else\newsavebox\spacPanelL\fi
\ifdefined\spacPanelR\else\newsavebox\spacPanelR\fi
\ifdefined\spacGapL\else\newlength\spacGapL\fi
\ifdefined\spacGapR\else\newlength\spacGapR\fi
\sbox\spacPanelL{%
\begin{minipage}[t]{0.485\textwidth}
\centering
\begin{tabular*}{\linewidth}[t]{@{\extracolsep{\fill}}llccccc@{}}
\toprule
\textbf{Method} & \textbf{Model} & \textbf{C1} & \textbf{C2} & \textbf{C3} & \textbf{C4} & \textbf{All} \\
\midrule
Mean & -- & \score{4.3} & \score{12.0} & \score{21.9} & \score{7.9} & \score{15.1} \\
Conversation Only & \textsc{GLM-5.3 Flash} & \score{41.7} & \score{39.0} & \score{22.0} & \score{23.2} & \score{27.6} \\
\midrule
\multicolumn{7}{@{}l@{}}{\textbf{In-Context Spatial Understanding}} \\
\multirow{2}{*}{Native Video} & \textsc{Gemini 3.7 Flash} & \score{74.3} & \score{55.9} & \score{31.5} & \score{37.3} & \score{41.4} \\
 & \textsc{Qwen3.8 27B} & \score{64.2} & \score{46.3} & \score{11.3} & \score{8.5} & \score{22.3} \\
\multirow{2}{*}{Uniform} & \textsc{GLM-5.3 Flash} & \score{38.3} & \score{43.3} & \score{20.5} & \score{16.8} & \score{26.2} \\
 & \textsc{GPT-5.6 Terra} & \score{63.6} & \score{54.8} & \score{29.3} & \score{33.1} & \score{38.5} \\
\multirow{2}{*}{AKS} & \textsc{GLM-5.3 Flash} & \score{41.5} & \score{44.9} & \score{21.0} & \score{18.2} & \score{27.3} \\
 & \textsc{GPT-5.6 Terra} & \score{64.5} & \score{55.5} & \score{28.2} & \score{34.4} & \score{38.5} \\
\multirow{2}{*}{FOCUS} & \textsc{GLM-5.3 Flash} & \score{38.8} & \score{43.0} & \score{20.0} & \score{17.3} & \score{26.1} \\
 & \textsc{GPT-5.6 Terra} & \score{59.0} & \score{55.9} & \score{28.9} & \score{34.8} & \score{38.7} \\
\bottomrule
\end{tabular*}
\end{minipage}
}%
\sbox\spacPanelR{%
\begin{minipage}[t]{0.485\textwidth}
\centering
\begin{tabular*}{\linewidth}[t]{@{\extracolsep{\fill}}llccccc@{}}
\toprule
\textbf{Method} & \textbf{Model} & \textbf{C1} & \textbf{C2} & \textbf{C3} & \textbf{C4} & \textbf{All} \\
\midrule
\multicolumn{7}{@{}l@{}}{\textbf{Agentic Spatial Understanding with Object Memory}} \\
VideoAgent & \textsc{GLM-5.3 Flash} & \score{36.2} & \score{38.7} & \score{16.3} & \na & \score{25.0} \\
Gemini Agentic & \textsc{Gemini 3.7 Flash} & \score{72.2} & \score{55.9} & \score{35.4} & \score{37.3} & \score{43.1} \\
\midrule
\multicolumn{7}{@{}l@{}}{\textbf{Multimodal RAG and Conversational Memory}} \\
\multirow{2}{*}{Multimodal RAG} & \textsc{GLM-5.3 Flash} & \score{41.7} & \score{23.0} & \score{4.1} & \score{2.4} & \score{10.6} \\
 & \textsc{GPT-5.6 Terra} & \score{35.3} & \score{26.0} & \score{5.3} & \score{3.9} & \score{11.8} \\
\multirow{2}{*}{Summary RAG} & \textsc{GLM-5.3 Flash} & \score{26.2} & \score{32.0} & \score{8.6} & \score{14.5} & \score{16.8} \\
 & \textsc{GPT-5.6 Terra} & \score{31.6} & \score{31.3} & \score{8.7} & \score{13.0} & \score{16.6} \\
\multirow{2}{*}{M2A} & \textsc{GLM-5.3 Flash} & \score{55.6} & \score{39.6} & \score{10.8} & \score{13.1} & \score{21.0} \\
 & \textsc{GPT-5.6 Terra} & \score{46.0} & \score{44.9} & \score{14.8} & \score{21.0} & \score{25.5} \\
\midrule
\multicolumn{7}{@{}l@{}}{\textbf{Ours}} \\
\multirow{2}{*}{SpaC-MEM} & \textsc{GLM-5.3 Flash} & \secondscore{85.0} & \bestscore{59.6} & \secondscore{38.2} & \secondscore{60.3} & \secondscore{51.5} \\
 & \textsc{GPT-5.6 Luna} & \bestscore{87.2} & \secondscore{59.0} & \bestscore{39.9} & \bestscore{61.4} & \bestscore{52.5} \\
\bottomrule
\end{tabular*}
\end{minipage}
}%
\setlength\spacGapL{0pt}\setlength\spacGapR{0pt}%
\ifdim\dimexpr\ht\spacPanelR+\dp\spacPanelR\relax>\dimexpr\ht\spacPanelL+\dp\spacPanelL\relax
  \setlength\spacGapL{\dimexpr(\ht\spacPanelR+\dp\spacPanelR-\ht\spacPanelL-\dp\spacPanelL)/11\relax}%
\else
  \setlength\spacGapR{\dimexpr(\ht\spacPanelL+\dp\spacPanelL-\ht\spacPanelR-\dp\spacPanelR)/13\relax}%
\fi
\begin{minipage}[t]{0.485\textwidth}
\centering
\begin{tabular*}{\linewidth}[t]{@{\extracolsep{\fill}}llccccc@{}}
\toprule
\textbf{Method} & \textbf{Model} & \textbf{C1} & \textbf{C2} & \textbf{C3} & \textbf{C4} & \textbf{All} \\
\midrule
Random & -- & \score{4.3} & \score{12.0} & \score{21.9} & \score{7.9} & \score{15.1} \\[\spacGapL]
Conversation Only & \textsc{GLM-5.3 Flash} & \score{41.7} & \score{39.0} & \score{22.0} & \score{23.2} & \score{27.6} \\[\spacGapL]
\midrule
\multicolumn{7}{@{}l@{}}{\textbf{In-Context Spatial Understanding}} \\[\spacGapL]
\multirow{2}{*}{Native Video} & \textsc{Gemini 3.7 Flash} & \score{74.3} & \score{55.9} & \score{31.5} & \score{37.3} & \score{41.4} \\[\spacGapL]
 & \textsc{Qwen3.8 27B} & \score{64.2} & \score{46.3} & \score{11.3} & \score{8.5} & \score{22.3} \\[\spacGapL]
\multirow{2}{*}{Uniform} & \textsc{GLM-5.3 Flash} & \score{38.3} & \score{43.3} & \score{20.5} & \score{16.8} & \score{26.2} \\[\spacGapL]
 & \textsc{GPT-5.6 Terra} & \score{63.6} & \score{54.8} & \score{29.3} & \score{33.1} & \score{38.5} \\[\spacGapL]
\multirow{2}{*}{AKS} & \textsc{GLM-5.3 Flash} & \score{41.5} & \score{44.9} & \score{21.0} & \score{18.2} & \score{27.3} \\[\spacGapL]
 & \textsc{GPT-5.6 Terra} & \score{64.5} & \score{55.5} & \score{28.2} & \score{34.4} & \score{38.5} \\[\spacGapL]
\multirow{2}{*}{FOCUS} & \textsc{GLM-5.3 Flash} & \score{38.8} & \score{43.0} & \score{20.0} & \score{17.3} & \score{26.1} \\[\spacGapL]
 & \textsc{GPT-5.6 Terra} & \score{59.0} & \score{55.9} & \score{28.9} & \score{34.8} & \score{38.7} \\[\spacGapL]
\bottomrule
\end{tabular*}
\end{minipage}
\hfill
\begin{minipage}[t]{0.485\textwidth}
\centering
\begin{tabular*}{\linewidth}[t]{@{\extracolsep{\fill}}llccccc@{}}
\toprule
\textbf{Method} & \textbf{Model} & \textbf{C1} & \textbf{C2} & \textbf{C3} & \textbf{C4} & \textbf{All} \\
\midrule
\multicolumn{7}{@{}l@{}}{\textbf{Agentic Spatial Understanding with Object Memory}} \\[\spacGapR]
VideoAgent & \textsc{GLM-5.3 Flash} & \score{36.2} & \score{38.7} & \score{16.3} & 13.4 & \score{21.0} \\[\spacGapR]
Gemini Agentic & \textsc{Gemini 3.7 Flash} & \score{72.2} & \score{55.9} & \score{35.4} & \score{37.3} & \score{43.1} \\[\spacGapR]
\midrule
\multicolumn{7}{@{}l@{}}{\textbf{Multimodal RAG and Conversational Memory}} \\[\spacGapR]
\multirow{2}{*}{Multimodal RAG} & \textsc{GLM-5.3 Flash} & \score{41.7} & \score{23.0} & \score{4.1} & \score{2.4} & \score{10.6} \\[\spacGapR]
 & \textsc{GPT-5.6 Terra} & \score{35.3} & \score{26.0} & \score{5.3} & \score{3.9} & \score{11.8} \\[\spacGapR]
\multirow{2}{*}{Summary RAG} & \textsc{GLM-5.3 Flash} & \score{26.2} & \score{32.0} & \score{8.6} & \score{14.5} & \score{16.8} \\[\spacGapR]
 & \textsc{GPT-5.6 Terra} & \score{31.6} & \score{31.3} & \score{8.7} & \score{13.0} & \score{16.6} \\[\spacGapR]
\multirow{2}{*}{M2A} & \textsc{GLM-5.3 Flash} & \score{55.6} & \score{39.6} & \score{10.8} & \score{13.1} & \score{21.0} \\[\spacGapR]
 & \textsc{GPT-5.6 Terra} & \score{46.0} & \score{44.9} & \score{14.8} & \score{21.0} & \score{25.5} \\[\spacGapR]
\midrule
\multicolumn{7}{@{}l@{}}{\textbf{Ours}} \\[\spacGapR]
\multirow{2}{*}{SpaC-MEM} & \textsc{GLM-5.3 Flash} & \secondscore{85.0} & \bestscore{59.6} & \secondscore{38.2} & \secondscore{60.3} & \secondscore{51.5} \\[\spacGapR]
 & \textsc{GPT-5.6 Luna} & \bestscore{87.2} & \secondscore{59.0} & \bestscore{39.9} & \bestscore{61.4} & \bestscore{52.5} \\[\spacGapR]
\bottomrule
\end{tabular*}
\end{minipage}
\end{table*}

\paragraph{RQ1.}
We first compare \methodshort{} with existing methods. As shown in \autoref{tab:main-results-no-ci}, \methodshort{} achieves the highest overall accuracy on \benchmark{} (52.5\%), outperforming the strongest baseline (43.1\%) using GPT-5.6 Luna as the answer model. The advantage becomes larger on more demanding spatial queries: \methodshort{} reaches 61.4\% on C4, compared with 37.3\% for the strongest baselines, while remaining competitive on the rest. \methodshort{} also maintains stronger accuracy as recall breadth and the number of prior frames before query increase. Importantly, it achieves this performance using only about one quarter of the reference input tokens required by native-video methods, showing that an explicit object-centric memory can preserve the information needed for \questionshort{} without repeatedly processing the full multimodal history.

\begin{figure}[t]
    \centering
    \includegraphics[width=0.9\linewidth]{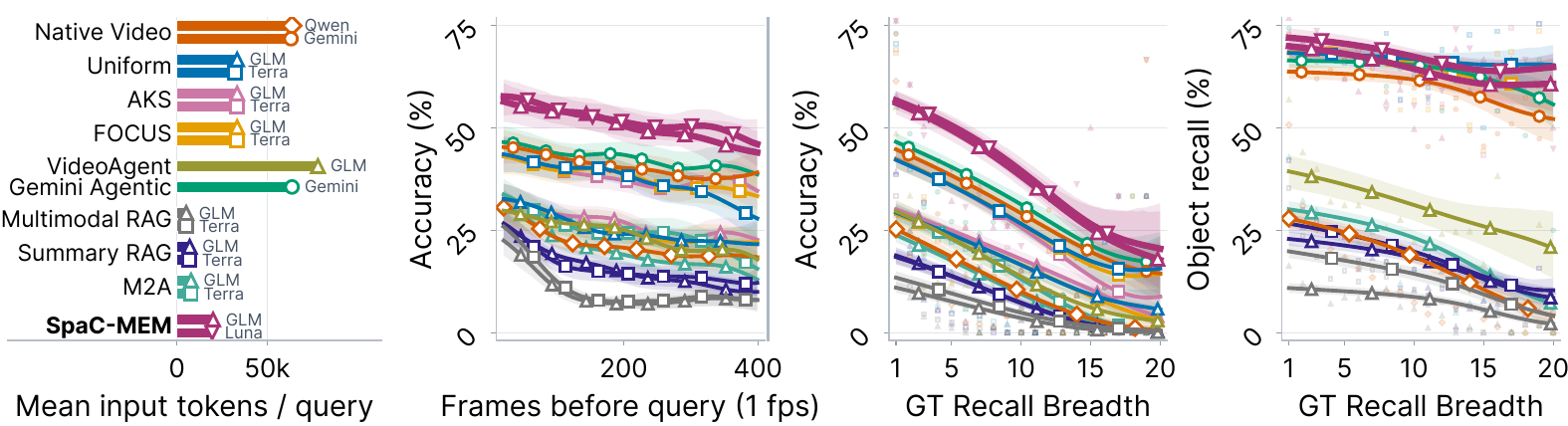 }
\caption{Left: reference input tokens across methods and accuracy as a function of input frames before queries. \methodshort{} achieves the highest accuracy while using substantially fewer reference input tokens. Right: object recall and answer accuracy across recall breadth. Colors and markers are consistent across panels; \methodshort{} maintains the high recall and accuracy as breadth increases.}
    \label{fig:token_recall}
\end{figure}

\begin{figure}
    \centering
    \includegraphics[width=0.9\linewidth]{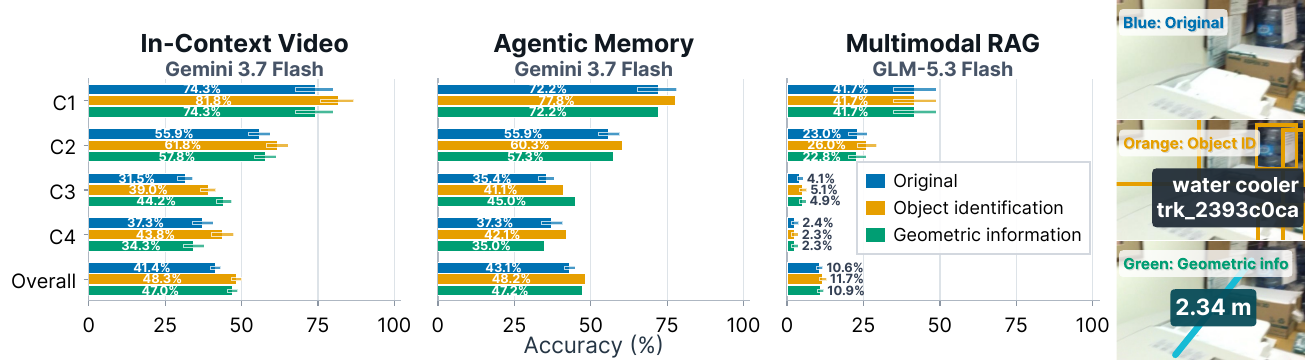}
\caption{Oracle ablations between the original frame, oracle object box, and oracle geometric measurement (illustrated on the right panel). Ground-truth boxes yield only marginal gains, while oracle geometric measurements mainly improve C3 but not C4.}
    \label{fig:oracle}
\end{figure}

\textbf{RQ2.}
To separate object-grounding errors from spatial-reasoning errors, we perform controlled oracle ablations for representative baselines in each method group. (1) \emph{Oracle object bounding boxes} provide ground-truth boxes to remove object localization and separation errors. As shown in \autoref{fig:oracle}, this yields only a marginal improvement, suggesting that object identification is not the primary bottleneck. (2) \emph{Oracle geometric measurements} additionally provide the spatial quantities required by each query. This improves in-context and agentic baselines on C3, which directly tests geometric reasoning, but produces little improvement on C4 counterfactual queries. These results suggest that the main limitation of existing baselines relative to \methodshort{} is not object detection or geometric measurement alone, but the lack of a consistent spatial representation that can preserve and combine spatial evidence across observations for \questionshort{}.


\begin{figure}
    \centering
    \includegraphics[width=0.9\linewidth]{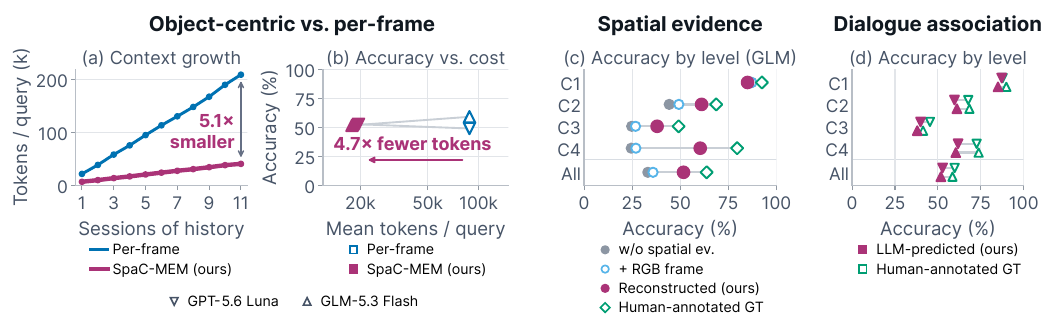}
\caption{Ablations of \methodshort{}: object-centric memory needs $4.7\times$ fewer tokens than per-frame memory (a, b), removing spatial evidence lowers accuracy (c), and LLM-predicted dialogue association stays close to human-annotated GT (d).}
    \label{fig:ablation}
\end{figure}
\label{sec:rq3}

\paragraph{RQ3.}
\autoref{fig:ablation} evaluates design choices in \methodshort{}. (1) \emph{Object-centric vs.\ per-frame memory:} \methodshort{} reduces context by $5.1\times$ at 11 sessions and uses $4.7\times$ fewer input tokens, with comparable accuracy. (2) \emph{Geometric vs.\ non-geometric memory:} removing spatial detail reduces accuracy from 51.5\% to 34.9\%, and C4 from 60.3\% to 24.5\%; adding each object's RGB view recovers little of this loss. The oracle reconstruction provides a further reference for the quality of the object records. (3) \emph{LLM vs.\ human dialogue association:} the method retains 89\% of the accuracy obtained with human-annotated associations, showing that most conversational grounding is preserved.


\section{Conclusion}
\label{sec:conclusion}

We introduced \benchmark{} to evaluate spatially grounded conversational reasoning across egocentric observations and evolving dialogue, and \methodshort{} to connect both histories through persistent object records. \methodshort{} achieves the highest overall answer accuracy among the evaluated methods while requiring fewer reference input tokens than native-video baselines. Ablations show that retaining object geometry supports both answer accuracy and object recall.

\label{maintextend}
\clearpage
\begingroup
\makeatletter
\let\@biblabel\NAT@biblabelnum
\let\@bibsetup\NAT@bibsetnum
\makeatother
\bibliography{references,references_arxiv_self}
\bibliographystyle{iclr2027_conference}
\endgroup
\clearpage
\appendix
\section{Related Work Comparison}
\label{app:related_work}

\begin{table}[!h]
\caption{Benchmark coverage of visual input, conversation, object identity, and evaluation. $\checkmark$: explicit support; \protect\LEFTcircle: related or limited support; $\times$: not specified.}
\label{tab:related}
\centering
\begingroup
\scriptsize
\setlength{\tabcolsep}{2.7pt}
\renewcommand{\arraystretch}{1.09}
\newcommand{\prophead}[1]{\rotatebox{90}{\shortstack[l]{#1}}}
\newcommand{\yesprop}{\ensuremath{\checkmark}}
\newcommand{\noprop}{\ensuremath{\times}}
\newcommand{\partialprop}{\LEFTcircle}
\newcommand{\datasetgroup}[1]{\multicolumn{10}{@{}l}{\textit{#1}}\\[2pt]}
\begin{tabularx}{\linewidth}{@{}>{\raggedright\arraybackslash}X*{9}{c}@{}}
\toprule
& \multicolumn{2}{c}{Visual input}
& \multicolumn{2}{c}{Conversation}
& Objects
& \multicolumn{4}{c}{Evaluation} \\
\cmidrule(lr){2-3}\cmidrule(lr){4-5}\cmidrule(lr){6-6}\cmidrule(lr){7-10}
& & & & &
& \multicolumn{3}{c}{Spatial reasoning}
& \\
\cmidrule(lr){7-9}
Benchmark
& \prophead{Egocentric\\video}
& \prophead{Real\\scenes}
& \prophead{Dialogue\\history}
& \prophead{Object-specific\\user semantics}
& \prophead{Cross-view \\ object continuity}
& \prophead{View\\grounding}
& \prophead{Spatial\\relations}
& \prophead{Geometric\\properties}
& \prophead{Recall breadth} \\
\midrule
\datasetgroup{First-person perception}
EPIC-KITCHENS \citep{damen2018epic}
& \yesprop & \yesprop & \noprop & \noprop & \noprop
& \partialprop & \noprop & \noprop & \noprop \\
Ego4D \citep{grauman2022ego4d}
& \yesprop & \yesprop & \noprop & \noprop & \yesprop
& \partialprop & \partialprop & \partialprop & \yesprop \\
\midrule
\datasetgroup{Situated dialogue and spatial understanding}
SIMMC-VR \citep{wu2023simmc}
& \yesprop & \noprop & \yesprop & \yesprop & \yesprop
& \yesprop & \yesprop & \noprop & \noprop \\
OpenEQA \citep{majumdar2024openeqa}
& \yesprop & \yesprop & \noprop & \noprop & \noprop
& \partialprop & \yesprop & \partialprop & \yesprop \\
SQA3D \citep{ma2023sqa3d}
& \yesprop$^{\dagger}$ & \yesprop & \noprop & \noprop & \noprop$^{\ddagger}$
& \partialprop & \yesprop & \noprop & \yesprop \\
VSI-Bench \citep{yang2025thinking}
& \yesprop & \yesprop & \noprop & \noprop & \noprop$^{\ddagger}$
& \partialprop & \yesprop & \yesprop & \yesprop \\
VSI-SUPER \citep{yang2025cambrian}
& \yesprop & \yesprop & \noprop & \noprop & \yesprop
& \partialprop & \noprop & \noprop & \yesprop \\
VSI-Super-Wild \citep{gu2026wild}
& \yesprop & \yesprop & \noprop & \noprop & \yesprop
& \partialprop & \partialprop & \noprop & \yesprop \\
\midrule
\datasetgroup{Video question answering}
EgoVQA \citep{fan2019egovqa}
& \yesprop & \yesprop & \noprop & \noprop & \noprop
& \partialprop & \partialprop & \noprop & \yesprop \\
EgoSchema \citep{mangalam2023egoschema}
& \yesprop & \yesprop & \noprop & \noprop & \noprop
& \noprop & \noprop & \noprop & \noprop \\
MULTIHOP-EGOQA \citep{chen2025multihop}
& \yesprop & \yesprop & \noprop & \noprop & \yesprop
& \partialprop & \partialprop & \noprop & \yesprop \\
LongVideoBench \citep{wu2024longvideo}
& \noprop & \yesprop & \noprop & \noprop & \yesprop
& \partialprop & \noprop & \noprop & \noprop \\
Video-MME \citep{fu2025videomme}
& \noprop & \yesprop & \noprop & \noprop & \noprop
& \partialprop & \partialprop & \noprop & \yesprop \\
\midrule
\datasetgroup{Conversational memory}
LoCoMo \citep{maharana2024locomo}
& \noprop & \noprop & \yesprop & \noprop & \noprop
& \noprop & \noprop & \noprop & \yesprop \\
LongMemEval \citep{wu2025longmemeval}
& \noprop & \noprop & \yesprop & \noprop & \noprop
& \noprop & \noprop & \noprop & \yesprop \\
\midrule
\textbf{\benchmark{} (ours)}
& \yesprop & \yesprop & \yesprop & \yesprop & \yesprop
& \yesprop & \yesprop & \yesprop & \yesprop \\
\bottomrule
\end{tabularx}
\par\vspace{3pt}
\begin{minipage}{\linewidth}
\scriptsize
$^{\dagger}$SQA3D includes an evaluated egocentric-video input option.
$^{\ddagger}$Static scene annotations alone are not treated as explicit cross-view object continuity.
\end{minipage}
\endgroup
\end{table}

\section{Additional Benchmark Details}
\label{app:benchmark}

\subsection{ScanNet observations and episode composition}
We build the visual history from ScanNet RGB-D recordings~\citep{dai2017scannet}. RGB frames form the observations shown to the assistant. Camera poses, frame-level instance annotations, and the annotated 3D mesh give us persistent object identities, visibility, and spatial measurements for constructing and checking questions. The instance labels link the egocentric views to the reference geometry, letting us ask about objects seen in video and check their spatial properties in a consistent physical scene.

Each conversation combines clips from selected scenes into several visits. The visits share a task and a pool of object-specific semantic facts $\Phi$, while recording gaps allow a conversation to continue in another room or return to an earlier one. We preserve object identity when a scene is revisited and keep objects from different scenes distinct. The rooms and their geometry remain those of the source scans; task roles, semantic facts, dialogue, and the order of visits are composed for the benchmark. For dataset splits, all clips from a physical scene stay in the same split. We use shards 01--19 of the v1.0 release: 95 conversations with 620 video sessions. Of 3,218 released questions, 3,100 enter the official evaluation after excluding 118 questions that do not require spatial evidence. Conversation statistics use the full histories; query statistics use this evaluation subset.

\subsection{Tasks, personas, and conversational state}
\label{app:tasks_personas}
\begin{table}[t]
\caption{Five everyday tasks used to guide the conversational facts in each episode. Each episode keeps one task and a compatible persona across all selected clips; Appendix~\ref{app:tasks_personas} describes their role in conversation generation.}
\label{tab:tasks}
\centering
\small
\setlength{\tabcolsep}{4.5pt}
\renewcommand{\arraystretch}{1.12}
\begin{tabularx}{\linewidth}{@{}L{0.18\linewidth}L{0.32\linewidth}YL{0.19\linewidth}@{}}
\toprule
Task & Purpose & Example properties $p$ & Example persona \\
\midrule
\textbf{Inventory walkthrough}
& Record what is in the space for later use.
& Owner, provenance, history, and alias.
& Roommate; new tenant. \\
\addlinespace[3pt]

\textbf{Guided room tour}
& Explain the space so another person can use it later.
& Function, history, owner, and alias.
& Welcoming host; prospective tenant. \\
\addlinespace[3pt]

\textbf{Lab or office setup}
& Prepare a shared work area without disturbing fixed items.
& Constraint, responsibility, maintenance, and priority.
& Lab member; organizer. \\
\addlinespace[3pt]

\textbf{Workspace setup}
& Decide how to arrange a usable workspace.
& Preference, intention, constraint, and function.
& Remote worker; interior designer. \\
\addlinespace[3pt]

\textbf{Cleaning and organizing}
& Decide what should be cleaned, kept, moved, or discarded.
& Maintenance, status, decision, and planning.
& Organizer; new tenant. \\
\bottomrule
\end{tabularx}
\end{table}

The five task settings give the conversation a reason to retain information about particular objects. An inventory can concern ownership and provenance; a room tour can establish functions and local names. Lab or office setup introduces responsibilities and access; workspace setup introduces intended uses and preferences; and cleaning and organizing distinguishes condition from a decision about what to do. We pair each task with a compatible persona and one of its three authored narrative variants. The role and way of speaking remain stable across visits. A task or persona frames the dialogue; an object acquires a semantic property only when the conversation establishes it.

We write an object-specific fact as $\phi=(o,p,x)$, with physical object $o$, property $p$, and value $x$, and retain when it was stated. The fact pool may include later updates. A \emph{correction} replaces an earlier value, including for retrospective questions; a \emph{state change} starts a new period while preserving the earlier value for questions about the past. The query program uses the value applicable to its temporal scope.

References are grounded in the observation and dialogue history available at the turn where they occur. A class name can identify an object when the context makes it unique. Otherwise, an appearance, spatial relation, established name, or clarification distinguishes it from similar objects. Descriptions tied to a particular view are used in the window where they were checked; an established name can carry the reference into later visits. This gives the Reader a recoverable link between conversational facts and ScanNet objects.

\subsection{Conversation generation}
\paragraph{Model versions.}
We use GPT-5.6 Sol as the Writer and GPT-5.6 Terra as the independent Reader, with low reasoning effort for the Reader. GPT-5.6 Luna resolves free-form object descriptions during evaluation.

\begin{table}[t]
\caption{Selected inputs to the Writer, with illustrative field values from one organizing task. Spatial context supports object references without disclosing the spatial outcomes being tested. Visual descriptions are checked independently before inclusion; query answers are never supplied.}
\label{tab:writer}
\centering
\begingroup
\small
\setlength{\tabcolsep}{0pt}
\renewcommand{\arraystretch}{1.03}
\begin{tabularx}{\linewidth}{@{}>{\raggedright\arraybackslash}X@{}}
\toprule
\textbf{Task and conversation context} \\
\textbf{Task and persona:} The continuing goal and the speaker's role and manner.\newline
\textit{Example: Cleaning and organizing; a considerate organizer; continue in the study.} \\
\textbf{Prior context:} Earlier dialogue, established object facts, and unfinished discussion.\newline
\textit{Example: The living room lamp was marked for cleaning; the walkthrough continues.} \\
\midrule
\textbf{Observation windows} \\
\textbf{Time and view:} A video interval and its representative frame.\newline
\textit{Example: Study window $w_1$, frames 120--180; representative frame 145.} \\
\textbf{Visibility changes:} Objects in view, newly encountered, revisited, or leaving view.\newline
\textit{Example: Two lamps and a cabinet appear; a previously seen sofa remains visible.} \\
\textbf{Spatial context:} Relations available in the window that can support object descriptions.\newline
\textit{Example: Cabinet near sofa, available in $w_1$ after both objects have been observed.} \\
\midrule
\textbf{Object references} \\
\textbf{Object identity:} The object's class, earlier sightings, and similar objects to distinguish.\newline
\textit{Example: Two study lamps are distinct from each other and from the living room lamp.} \\
\textbf{Permitted descriptions:} Natural references and the observations needed to use them.\newline
\textit{Example: ``The cabinet near the sofa'' becomes usable after both objects are observed.} \\
\midrule
\textbf{Visual references for ambiguous objects} \\
\textbf{Checked description:} A VLM describes a marked object; an independent VLM, without the target cue, must identify only that object from the phrase.\newline
\textit{Example: ``The lamp with the red shade'' uniquely resolves to the intended study lamp.} \\
\textbf{Valid frame:} The exact view where the phrase was verified; later references require an established name or another valid description.\newline
\textit{Example: Use the phrase at frame 145; name it ``reading lamp'' in that turn for later references.} \\
\midrule
\textbf{Fact pool} \\
\textbf{Required facts:} Object properties that must be expressed before their dependent questions.\newline
\textit{Example: Reading lamp: maintenance = needs cleaning.} \\
\textbf{Optional facts:} Additional task information the Writer may include naturally.\newline
\textit{Example: Cabinet: comment = cluttered.} \\
\textbf{Timing and dependencies:} When a fact can be introduced and which statements must precede it.\newline
\textit{Example: Establish the lamp's cleaning fact in $w_1$ before the question in $w_2$.} \\
\midrule
\textbf{Question intents} \\
\textbf{Meaning and answer type:} The conversational condition, spatial test, and requested result.\newline
\textit{Example: Marked for cleaning; on a table; return the number of matching objects.} \\
\textbf{Scope and placement:} Which observations the question covers and when it may be asked.\newline
\textit{Example: All rooms visited so far; ask in $w_2$ after all required facts are established.} \\
\bottomrule
\end{tabularx}
\endgroup
\end{table}

We select queries from $\Phi$ whose semantic prerequisites and spatial evidence can be available before the question. The selected queries cover the complexity classes and recall breadths defined in the main text. Each has a deterministic program $G_k$, a temporal scope, and a placement window in which its answer is stable. Required facts are scheduled before that window. Other compatible facts can enter the dialogue as the task develops.

An LLM \emph{Writer} generates the conversation in scene-level batches. It receives the task and persona, the preceding dialogue and summary, relevant RGB observation windows, grounded reference options, and the facts to establish. The query specification tells it what condition to ask about and when; the answer and qualifying object set are withheld. Carrying the prior dialogue across batches lets a later visit continue the same task and refer back to established facts. \autoref{tab:writer} summarizes the principal inputs.

An independent \emph{Reader} then processes the realized conversation turn by turn. Using the visual and dialogue history available at each turn, it resolves mentions and reconstructs the stated facts and updates. We use this recovered state for answer execution, so the benchmark answer follows the conversation that was written.

\subsection{Query construction and recall}
The four spatial complexity classes follow \autoref{sec:questions}. Qualitative and geometric predicates operate on objects and measurements from the same source scene, even when a question aggregates results across scenes. Counterfactual predicates use objects already observed and evaluate a hypothetical viewpoint or placement against the available scene geometry. Recall-oriented questions require the complete set of objects satisfying both the conversational and spatial conditions. We retain that set alongside the final answer so a correct count can be checked against the objects that support it.

\subsection{Ground truth and answer evaluation}
\label{sec:truth}
Let $\Phi_{<k}$ be the Reader-recovered facts established before $q_k$, resolved for its temporal scope, and $\mathcal G_{\leq t_k}$ the ScanNet spatial evidence available up to $t_k$. We compute the canonical answer by executing the question program over that prefix:
\begin{equation}
a_k=\operatorname{Exec}(G_k;\Phi_{<k},\mathcal G_{\leq t_k}).
\label{eq:replay}
\end{equation}
Execution also records the qualifying objects and the spatial measurements that support the answer. This matters for a count: two methods can return the same integer while identifying different objects.

For count questions involving measured geometric boundaries, we retain the literal comparison in $G_k$ and record how far each candidate lies from the boundary. Candidates clearly inside contribute to the lower count $L_k$; those within the measurement tolerance also contribute to the upper count $U_k$. The accepted answer is any integer in $[L_k,U_k]$. We use a $0.04$\,m tolerance for distance and extent comparisons, reflecting the resolution of the annotated ScanNet mesh. During query selection, we bound the interval width to two counts. When no candidate lies near the boundary, $L_k=U_k=a_k$. This treatment preserves the stated spatial condition while allowing for small measurement differences at its edge.

Models still return a single count. We score it by membership in the accepted interval when one applies; other count answers and Boolean answers use exact matching. For object identification, a VLM maps the model's natural-language description to a physical object using the observations available before the question. The mapped object is compared with the target fixed by $G_k$. Ambiguous descriptions remain unresolved. Models see neither the hidden object identifiers nor a candidate answer list.

\subsection{Verification and quality review}
We check each realized question against the Reader-recovered prefix. Its required facts must have been stated before the question, its references must resolve at the turns where they are used, and its objects and spatial evidence must be available by the query time. We execute the program again at the realized placement and retain the question when the result agrees with the planned answer. This also checks that conversational updates and optional facts have preserved the intended temporal and semantic conditions. Visual descriptions used for object answers must provide enough evidence for reliable grounding.

Human review examines the resulting video and dialogue for semantic grounding, reference resolution, temporal scope, conversational plausibility, and answer correctness. The automated checks and this review address different failure modes: execution checks the formal conditions, while the review checks that the question and conversation express them clearly.

\section{Control Baselines}
\label{app:controls}

\paragraph{Random.}
For count and Boolean questions, this control returns the mean canonical answer over the evaluated set, rounded to the nearest valid integer or Boolean value. For object identification, it samples uniformly from the distinct objects in the available memory before the query. Sampling uses seed 20260925 with a separate deterministic draw for each query. Count and Boolean predictions follow the standard answer rules; object predictions are compared directly with the target identity.

\paragraph{Conversation Only.}
We give GLM-5.3 Flash the timestamped public dialogue strictly before the query, followed by the question and its answer format. The system prompt and output format match the frame-based baselines, with all images and the frame-index block removed. We use low reasoning effort and a maximum output length of 32,768 tokens.

\section{Reference Input Cost}
\label{app:token_cost}

Provider-reported input-token counts are not directly comparable across our baselines because different models use different text tokenizers and visual tokenization schemes. We therefore report \emph{reference input tokens}, a model-independent measure of the context delivered to the final answer model. The goal is not to reproduce provider billing, but to compare how much textual and visual context each method requires at query time.

For method $m$ answering query $q$, we define
\[
T_{\mathrm{ref}}(m,q)
=
\left|
\operatorname{tok}_{\texttt{o200k}}
\!\left(s_m \Vert u_{m,q}\right)
\right|
+
c_{\mathrm{frame}}F_{m,q},
\]
where $s_m$ and $u_{m,q}$ are the system and user prompts actually provided to the answer model, and $F_{m,q}$ is the number of RGB frames provided with that request. We tokenize all text using the same \texttt{o200k\_base} tokenizer. All visual inputs are evaluated at $640\times480$ resolution and assigned a common reference cost of one token per $32\times32$ patch, giving $c_{\mathrm{frame}}=\lceil640/32\rceil\lceil480/32\rceil=300$ reference tokens per frame. For image-based methods, $F_{m,q}$ is the number of attached images; for native-video methods, it is the number of frames sampled into the causal video input at 1 fps.

This measure is computed offline from saved answer requests and is applied identically across methods. It measures the context exposed to the final answer model rather than total pipeline compute: retrieval embeddings, summary generation, memory updates, agent tool calls, output tokens, and evaluation calls are excluded. Internal frame re-inspection by agentic methods is likewise not charged again because it does not change the context of the final answer request. For text-only runs whose complete prompts were not retained, we use the recorded input-token count only when the serving tokenizer matches \texttt{o200k\_base}. We report the mean $T_{\mathrm{ref}}$ over all scored answer requests for each method.

\section{Additional Method Details}
\label{app:method}

\subsection{Object reconstruction and consolidation}
\label{app:method_reconstruction}

\paragraph{Geometry.}
We use the DA3-Nested-Giant-Large checkpoint of Depth Anything 3~\citep{lin2025depth} as $f_{\mathrm{geo}}$, yielding approximately metric geometry.
\methodshort{} places all frames of a scene in one shared coordinate system, with camera intrinsics, poses and depth all estimated from the RGB frames alone. Only pixels whose depth in $\hat{D}_t$ is estimated with high confidence are lifted into 3D. The scene's axes are aligned with its dominant planes: the vertical axis with the floor plane or, when too little floor is visible, with walls facing different directions or the largest horizontal surface, and the horizontal axes with the dominant wall direction, so that axis-aligned boxes follow the room's layout. The floor is placed at height zero.

\paragraph{Segmentation and lifting.} Wall and floor are excluded from $\mathcal{L}$, since they form the scene's structure rather than objects. Instances are kept when $f_{\mathrm{seg}}$ is sufficiently confident, which keeps spurious detections out of the memory. Where the model also ranks alternative classes, the next most likely ones are kept as runner-up classes. To decide which mask pixels are consistent with the object, \methodshort{} uses the mask's interior, away from its boundary, as a reference: $\tilde{M}_{t,i}$ keeps the pixels of $M_{t,i}$ whose depth lies within the interior's central depth range, widened by a small tolerance. Edge pixels whose depth matches the object are therefore kept, including thin parts such as chair legs. The centroid $\boldsymbol{\mu}_{t,i}$ is the mean of $\hat{P}_{t,i}$, and the box $\mathbf{B}_{t,i}$ spans $\hat{P}_{t,i}$ along each axis of the scene's coordinate system, excluding the most extreme few points, which makes it robust to stray points.

\paragraph{Observation matching.} Each observation $\hat{o}_{t,i}$ is compared with the objects in the scene's pool that carry its class label. Their overlap $A(\hat{o}_{t,i},\hat{o})$ is the share of the points of $\hat{o}_{t,i}$ that lie within 10\,cm of the accumulated points of $\hat{o}$. The match $a_t(\hat{o}_{t,i})$ is accepted when this share is large enough and clearly exceeds that of every other candidate. All observations of a frame are matched jointly, and each object takes at most one of them, since an object appears only once in a view. This keeps neighbouring objects of the same class apart even when they are seen together. A matched observation adds its points to $\hat{o}$, updating $s_{\leq t}(\hat{o})$. The class label of $\hat{o}$ is the one best supported across its observations, and other well-supported classes are kept as runner-up classes.

\paragraph{Pool refinement.} Two objects in the pool are merged when their 3D points overlap substantially, they never co-occur in the same frame, and they either share a track from $f_{\mathrm{seg}}$ or have similar appearance. Overlapping pieces that never co-occur are likely views of one object, and the shared track or appearance keeps distinct objects that happen to overlap apart. Each object is kept once it is confirmed by several observations. Isolated points are removed before the final box is computed, so that each box follows the object's main body.

\paragraph{Class labels and runner-up classes.}
Each persistent object carries one class label together with up to two runner-up classes, retained as secondary label evidence. Both the association model and the answering agent are told that these describe a single persistent object rather than additional objects, and that an object carrying a queried class label outright is preferred over one listing it as a runner-up. An object with no credible runner-up carries none.

\paragraph{Coordinates.}
Each scene retains its own levelled coordinate frame with approximately metric scale. Spatial predicates are evaluated within a scene, and cross-scene queries aggregate their results, so a shared origin and orientation across scenes are unnecessary.

\subsection{Dialogue association and working memory}
\label{app:method_memory}

\paragraph{Association model.}
For each query, $f_{\mathrm{assoc}}$ receives the preceding dialogue and objects observed by the query time, grouped by scene. Each object has an identifier, class labels, centroid and bounding box. Turns are supplied in order with objects visible within two seconds, restricted to the available query prefix; the query itself is withheld. The model resolves class descriptions, pronouns and established names using this context. It returns a turn, object, referring phrase and role for each association, with identifiers constrained to the supplied inputs. Only \emph{subject} roles enter $\hat{\mathcal{A}}_k$: in \emph{``the monitor beside the laptop is Sam's,''} the monitor is the subject and the laptop is a \emph{locator}. A unique class mention can help resolve a reference but does not establish a subject role by itself. References without an identifiable subject remain unassigned.

\paragraph{Contents of the working memory.}
$\mathcal{W}_k$ holds one record per persistent object observed up to $t_k$, grouped by scene and session and ordered by class label within a scene. Each record carries the object's class label and runner-up classes, its centroid and axis-aligned bounding box, the frame at which it first came into view, and the camera position at that frame, which the egocentric queries use as the position the user was standing at. Beneath the record sit the dialogue turns associated with the object, verbatim and in conversational order, each annotated so that the phrase referring to this object is marked. A turn about several objects therefore appears under each of them with a different phrase marked, and the answering agent is instructed to read the clause carrying the mark as the statement about that object. A table mapping days to scenes precedes the records, and $\hat{\mathcal{O}}_{t_k}$ follows them.

\section{VLM Judge for Object Identification}
\label{app:judge}

Object-identification queries require a method to name a single physical object in free-form text, for example, ``Which single thing I said was an early priority was near a copier?'' Correctness must therefore be determined by resolving the generated description to the ground-truth object. A target-only judge that sees only evidence for the correct object is too permissive in this setting: a generic answer such as ``the table'' may be accepted whenever the target is the only table in the displayed frame, even when several tables satisfy the query's conversational condition. Such a judge can therefore reward class-level guessing without establishing that the method resolved the spatial condition.

\paragraph{Candidate-resolution procedure.}
For each identification query, we first construct the candidate set from objects satisfying its conversation-derived condition. This set is fixed by the benchmark program before method evaluation and always contains the ground-truth target. Every candidate is then presented uniformly: one RGB frame is selected where its ground-truth instance mask is largest, only that object is boxed and assigned a letter, and candidates are shuffled with a fixed per-query seed. The judge is not told which candidate is correct.

Each candidate is accompanied by its object class, conversation facts, first-seen day, and a compact scene graph computed from ground-truth geometry. The graph contains objects within 3\,m of the centroid, ordered nearest first, together with \emph{on} and \emph{supports} relations; the benchmark definitions use \emph{near} $\leq 1.5$\,m and \emph{far} $\geq 3$\,m. This context allows descriptions such as ``the pillow by the whiteboard'' to be checked even when the referenced neighbour is outside the selected frame.

Given a model answer, the judge returns every candidate consistent with the answer's own description using object class, appearance, dialogue names and facts, and location claims. To avoid solving the query on behalf of the evaluated method, the judge does not apply the query's selection rule itself. Thus, a class-only answer or a restatement of the conversational condition remains compatible with every otherwise matching candidate. The judge also ignores location claims that merely repeat the query's own reference object; only additional location evidence supplied by the answer may eliminate a candidate, and only when that claim is clearly inconsistent with the candidate's scene graph.

An answer is correct if and only if the resolved candidate set contains exactly the ground-truth target. We use three independent judge calls at temperature 0 and take the majority verdict. Count and Boolean queries remain scored deterministically; the candidate-resolution judge is used only for object-identification queries.

To validate the judge, 500 identification answers sampled across 12 baselines were independently verified by a human annotator using the same candidate presentation. 
\newpage
\section{Additional Experimental Results}
\label{app:result}

\begin{table*}[t]
\centering
\scriptsize
\setlength{\tabcolsep}{1pt}
\renewcommand{\arraystretch}{1.18}
\newcommand{\score}[2]{%
  \begin{tabular}[t]{@{}c@{}}#1\\[-2.5pt]\textcolor{gray}{\tiny $\pm#2$}\end{tabular}}
\newcommand{\bestscore}[2]{%
  \begin{tabular}[t]{@{}c@{}}\textbf{#1}\\[-2.5pt]\textcolor{gray}{\tiny $\pm#2$}\end{tabular}}
\newcommand{\secondscore}[2]{%
  \begin{tabular}[t]{@{}c@{}}\underline{#1}\\[-2.5pt]\textcolor{gray}{\tiny $\pm#2$}\end{tabular}}
\newcommand{\na}{\textcolor{gray}{--}}
\providecommand{\SpaC}{SpaC}
\caption{Answer Accuracy across selected baselines and \SpaC-MEM (\%) with Wilson 95\% confidence interval half-widths in gray; bold and underlined entries are the best and second-best values per column.}
\label{tab:main-results}
\begin{tabular*}{\textwidth}{@{\extracolsep{\fill}}llccccc@{}}
\toprule
\textbf{Method} & \textbf{Model} & \textbf{C1} & \textbf{C2} & \textbf{C3} & \textbf{C4} & \textbf{All} \\
\midrule
Random & -- & \score{4.3}{3.0} & \score{12.0}{2.3} & \score{21.9}{2.1} & \score{7.9}{1.9} & \score{15.1}{1.3} \\
Conversation Only & \textsc{GLM-5.3 Flash} & \score{41.7}{7.0} & \score{39.0}{3.5} & \score{22.0}{2.2} & \score{23.2}{3.0} & \score{27.6}{1.6} \\
\midrule
\multicolumn{7}{@{}l@{}}{\textbf{In-Context Spatial Understanding}} \\
\multirow{2}{*}{Native Video} & \textsc{Gemini 3.7 Flash} & \score{74.3}{6.2} & \score{55.9}{3.5} & \score{31.5}{2.4} & \score{37.3}{3.5} & \score{41.4}{1.7} \\
 & \textsc{Qwen3.8 27B} & \score{64.2}{6.8} & \score{46.3}{3.6} & \score{11.3}{1.6} & \score{8.5}{2.0} & \score{22.3}{1.5} \\
\multirow{2}{*}{Uniform} & \textsc{GLM-5.3 Flash} & \score{38.3}{7.0} & \score{43.3}{3.6} & \score{20.5}{2.1} & \score{16.8}{2.7} & \score{26.2}{1.6} \\
 & \textsc{GPT-5.6 Terra} & \score{63.6}{6.8} & \score{54.8}{3.5} & \score{29.3}{2.4} & \score{33.1}{3.4} & \score{38.5}{1.7} \\
\multirow{2}{*}{AKS} & \textsc{GLM-5.3 Flash} & \score{41.5}{7.1} & \score{44.9}{3.6} & \score{21.0}{2.1} & \score{18.2}{2.8} & \score{27.3}{1.6} \\
 & \textsc{GPT-5.6 Terra} & \score{64.5}{6.9} & \score{55.5}{3.6} & \score{28.2}{2.4} & \score{34.4}{3.4} & \score{38.5}{1.7} \\
\multirow{2}{*}{FOCUS} & \textsc{GLM-5.3 Flash} & \score{38.8}{7.0} & \score{43.0}{3.6} & \score{20.0}{2.1} & \score{17.3}{2.7} & \score{26.1}{1.6} \\
 & \textsc{GPT-5.6 Terra} & \score{59.0}{7.1} & \score{55.9}{3.6} & \score{28.9}{2.4} & \score{34.8}{3.4} & \score{38.7}{1.7} \\
\midrule
\multicolumn{7}{@{}l@{}}{\textbf{Agentic Spatial Understanding with Object Memory}} \\
VideoAgent & \textsc{GLM-5.3 Flash} & \score{36.2}{7.1} & \score{38.7}{3.6} & \score{16.3}{2.0} & \score{13.4}{1.8} & \score{21.0}{1.8} \\
Gemini Agentic & \textsc{Gemini 3.7 Flash} & \score{72.2}{6.4} & \score{55.9}{3.5} & \score{35.4}{2.5} & \score{37.3}{3.5} & \score{43.1}{1.7} \\
\midrule
\multicolumn{7}{@{}l@{}}{\textbf{Multimodal RAG and Conversational Memory}} \\
\multirow{2}{*}{Multimodal RAG} & \textsc{GLM-5.3 Flash} & \score{41.7}{7.0} & \score{23.0}{3.0} & \score{4.1}{1.0} & \score{2.4}{1.1} & \score{10.6}{1.1} \\
 & \textsc{GPT-5.6 Terra} & \score{35.3}{6.8} & \score{26.0}{3.1} & \score{5.3}{1.2} & \score{3.9}{1.4} & \score{11.8}{1.1} \\
\multirow{2}{*}{Summary RAG} & \textsc{GLM-5.3 Flash} & \score{26.2}{6.3} & \score{32.0}{3.3} & \score{8.6}{1.5} & \score{14.5}{2.5} & \score{16.8}{1.3} \\
 & \textsc{GPT-5.6 Terra} & \score{31.6}{6.6} & \score{31.3}{3.3} & \score{8.7}{1.5} & \score{13.0}{2.4} & \score{16.6}{1.3} \\
\multirow{2}{*}{M2A} & \textsc{GLM-5.3 Flash} & \score{55.6}{7.0} & \score{39.6}{3.5} & \score{10.8}{1.6} & \score{13.1}{2.4} & \score{21.0}{1.4} \\
 & \textsc{GPT-5.6 Terra} & \score{46.0}{7.1} & \score{44.9}{3.5} & \score{14.8}{1.8} & \score{21.0}{2.9} & \score{25.5}{1.5} \\
\midrule
\multicolumn{7}{@{}l@{}}{\textbf{Ours}} \\
\multirow{2}{*}{SpaC-MEM} & \textsc{GLM-5.3 Flash} & \secondscore{85.0}{5.1} & \bestscore{59.6}{3.5} & \secondscore{38.2}{2.5} & \secondscore{60.3}{3.5} & \secondscore{51.5}{1.8} \\
 & \textsc{GPT-5.6 Luna} & \bestscore{87.2}{4.8} & \secondscore{59.0}{3.5} & \bestscore{39.9}{2.5} & \bestscore{61.4}{3.5} & \bestscore{52.5}{1.8} \\
\bottomrule
\end{tabular*}
\end{table*}

\newpage
\section{Additional Analysis of Object-Centric Memory}
\label{app:ablation}

\begin{figure}[t]
    \centering
    \includegraphics[width=\linewidth]{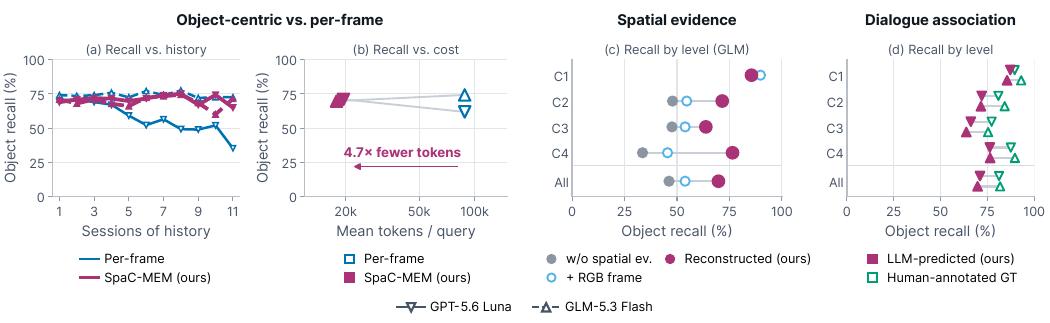}
    \caption{Object recall for the ablations in \autoref{fig:ablation}: \methodshort{} keeps about 70\% recall as history grows while per-frame memory degrades (a) at $4.7\times$ the token cost (b), spatial evidence drives recall on spatial queries (c), and LLM-predicted dialogue association retains most of the recall of human-annotated GT association (d).}
    \label{fig:ablation_recall}
\end{figure}

\subsection{Object recall in the ablations}
\autoref{fig:ablation_recall} reports the fraction of ground-truth objects recovered by the answering model. \emph{Object-centric vs.\ per-frame memory:} \methodshort{} maintains roughly 70\% recall as interaction history grows, while using $4.7\times$ fewer input tokens; with GPT-5.6 Luna, per-frame recall drops from 73\% in the first session to 35\% after 11 sessions. \emph{Geometric vs.\ non-geometric memory:} preserving 3D spatial records raises recall from 46.2\% to 69.7\% overall and from 33.6\% to 76.5\% on C4, showing that spatial evidence is critical for recovering the correct object set. \emph{LLM vs.\ human dialogue association:} automatic associations retain 87\% and 85\% of the recall achieved with human annotations for Luna and GLM-5.3 Flash, respectively.

\subsection{Scaling with memory size}
\begin{figure}[!h]
    \centering
    \includegraphics[width=\linewidth]{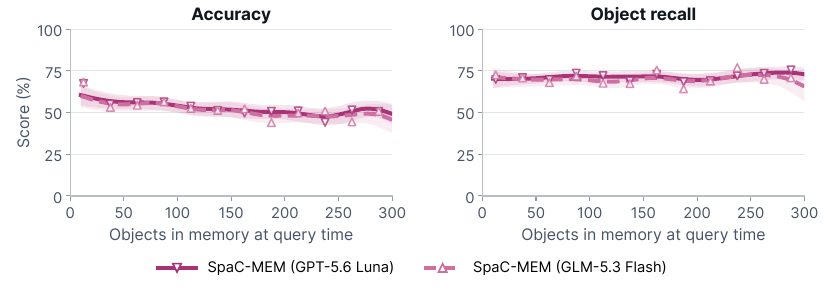}
    \caption{Answer accuracy and object recall as the number of objects in $\mathcal{W}_k$ grows. Solid and dashed lines use GPT-5.6 Luna and GLM-5.3 Flash, respectively. Curves are smoothed trends with 95\% episode-bootstrap intervals; markers show means within 25-object bins. The horizontal axis covers 95\% of queries.}
    \label{fig:memory_size}
\end{figure}

\autoref{fig:memory_size} relates answer accuracy and object recall to the number of persistent objects in the working memory $\mathcal{W}_k$ at query time. Across the range covering 95\% of queries, object recall remains near 70\% for both answering models, while accuracy changes gradually with memory size. In the object-centric design of \autoref{sec:method}, repeated observations update the same object record, and its associated dialogue remains available to the answerer.

\subsection{Fidelity of reconstructed objects}
\label{app:recon_vs_gt}
The object reconstruction and consolidation stages build the persistent pool $\hat{\mathcal{O}}_{\leq t}$ from RGB observations. We compare its spatial records $s_{\leq t}(\hat{o})$ with annotated ScanNet objects to assess how much object evidence reaches the working memory. Each of the 39 benchmark scenes is evaluated once, giving 864 annotated objects in the ScanNet200 vocabulary. We align the reconstructed and ScanNet coordinate systems using corresponding camera positions, then match objects one-to-one by centroid distance. Matching is independent of class; semantic recall additionally requires the reconstructed class label to agree with the annotation.

\begin{figure}[t]
\centering
\includegraphics[width=\linewidth]{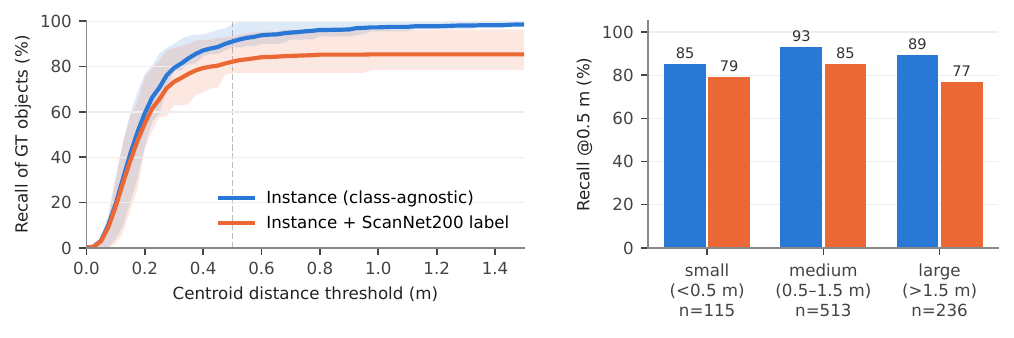}
\caption{Reconstruction recall by centroid-distance threshold (left) and annotated object size at 0.5\,m (right). The bands show the per-scene interquartile range. Instance recall is class-agnostic; semantic recall also requires the correct ScanNet200 label.}
\label{fig:recon_vs_gt}
\end{figure}

\autoref{tab:recon_vs_gt} reports object-level results at three distance thresholds. At 0.5\,m, the persistent pool recovers 91.1\% of annotated objects, and 82.2\% are recovered with the correct class label. The median centroid error among matches is 16.2\,cm. \autoref{fig:recon_vs_gt} shows how recall varies with matching distance and object size.

\begin{table}[t]
\centering
\small
\setlength{\tabcolsep}{5pt}
\begin{tabular}{lccc}
\toprule
 & 0.25\,m & 0.5\,m & 1.0\,m \\
\midrule
Instance recall (\%) & 70.7 & \textbf{91.1} & 97.3 \\
Semantic recall (\%) & 65.6 & \textbf{82.2} & 85.4 \\
Instance precision (\%) & 52.0 & 67.0 & 71.6 \\
\midrule
\multicolumn{4}{l}{Median centroid error of matched objects: 16.2\,cm} \\
\multicolumn{4}{l}{Recall at 3D box IoU 0.25: 51.7\%} \\
\bottomrule
\end{tabular}
\caption{Reconstructed object records compared with ScanNet annotations across 39 scenes and 864 annotated objects. Instance matching uses centroid distance; semantic recall also requires agreement on the ScanNet200 class.}
\label{tab:recon_vs_gt}
\end{table}

\end{document}